\documentclass[11pt]{article}
\pdfoutput=1 

\usepackage[preprint]{acl}

\usepackage{times}
\usepackage{latexsym}

\usepackage[T1]{fontenc}

\usepackage[utf8]{inputenc}

\usepackage{microtype}

\usepackage{inconsolata}

\usepackage{graphicx}
\usepackage{subcaption}
\usepackage{xspace}
\usepackage[table]{xcolor}
\usepackage{tcolorbox}
\definecolor{darkred}{RGB}{160, 0, 0}
\usepackage{booktabs}
\usepackage{multirow}
\usepackage{amsfonts}
\newcommand{\name}[0]{{Koa-decision}\xspace}
\usepackage{amsmath}
\usepackage{hyperref}
\usepackage{cleveref}

\title{Koa-decision: Fast and Consistent Structured Decision Making with Generative LLMs\thanks{An earlier version of this work was made available on \href{https://openreview.net/forum?id=78WdKlYSeO}{OpenReview} in September 2025}}

\author{
  \textbf{Shenghong Dai\textsuperscript{1,2}},
  \textbf{Shiva Kumar Pentyala\textsuperscript{1}},
  \textbf{Yingchi Liu\textsuperscript{1}},
  \textbf{Shubham Mehrotra\textsuperscript{1}},
\\
  \textbf{Suman Banerjee\textsuperscript{2}},
  \textbf{James Zhu\textsuperscript{1}},
  \textbf{Bin Bi\textsuperscript{1}},
  \textbf{Sitaram Asur\textsuperscript{1}},
  \textbf{Phil Mui\textsuperscript{1}}
\\
\\
  \textsuperscript{1}Salesforce AI,
  \textsuperscript{2}University of Wisconsin--Madison
\\
}

\begin{document}
\maketitle
\begin{abstract}
Industry applications often demand low-latency classification, yet current large language model (LLM) approaches remain poorly suited for latency-critical applications. Existing prompting and constrained decoding produce verbose, multi-token outputs that require expensive token-by-token generation, while encoder-based models achieve faster inference but sacrifice the task flexibility. We propose \name, a framework for low-latency \emph{atomic actions}---fast, single-step decisions such as classification, semantic endpointing, Boolean checks, and scoring---formulated as constrained generation with single-token outputs. By introducing atomic label tokens and applying supervised fine-tuning, our method reduces classification to a deterministic one-step decoding problem. Across standard benchmarks, \name delivers competitive accuracy with consistently low and stable latency. On a \textbf{production intent-routing benchmark}, \name reaches \textbf{85.5\%} accuracy---competitive with the strongest frontier models (Claude-4.8-Opus, Gemini-Pro-3.1) and ahead of GPT-5 and Gemini-2.5-Pro---while answering in about \textbf{half a second}---several-fold faster than every frontier model (up to $\sim$7.5$\times$ at the median) under identical serving conditions. Against the dedicated single-token system Jev/TypeSafe, \name is competitive on accuracy and faster at the median, while also handling multimodal inputs and multi-label outputs that single-label text systems do not.
\end{abstract}

\section{Introduction}

Industry applications rely heavily on classification for a wide range of decisions, from sentiment analysis~\citep{pang2008opinion} and intent recognition~\citep{goo2018slot,chen2019bert} to customer support and interactive dialogue agents. As these applications increasingly operate on multimodal data---combining text and vision---there is growing demand for unified models that can handle diverse input modalities while maintaining efficiency in latency-sensitive environments~\citep{wang2024qwen2}.

Current approaches to classification with large language models (LLMs) face significant limitations, particularly in latency-critical applications. Prompt-based methods, while intuitive, often produce verbose, multi-token responses that require additional parsing, which introduces substantial inference overhead. More importantly, they provide no guarantee that outputs will be single tokens: a request such as ``classify this review as positive or negative'' can yield explanatory sentences or multi-token paraphrases rather than clean categorical labels. Even with constrained decoding techniques \citep{geng2023grammar} that restrict outputs to valid label strings, models still rely on token-by-token generation. This scales poorly with label vocabulary size and leads to unpredictable latency variations.

This latency challenge is particularly acute in real-world deployment scenarios where classification must occur at scale with strict response time requirements. Traditional encoder-based approaches (e.g., BERT with classification heads) offer predictable, low-latency inference but lack the flexibility and generalization capabilities that make modern LLMs attractive for complex reasoning tasks, since they require task-specific architectures and dataset-specific fine-tuning.

\begin{figure}[t]
    \centering
    \includegraphics[width=0.8\linewidth]{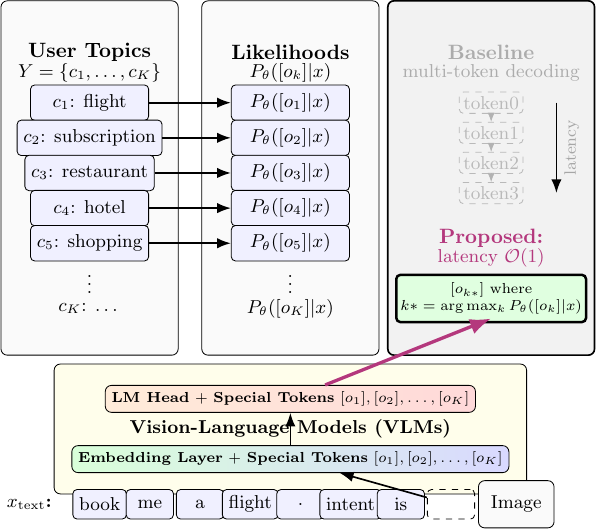}
    \caption{
    \textbf{Overview of our framework.} 
     Inputs are processed by an LLM that directly outputs an atomic special token for the target class. Unlike baseline prompting with multi-token decoding, the framework performs classification in a single $\mathcal{O}(1)$ decode step (independent of output length) and supports zero-shot adaptation.
    }
    \label{fig:main_idea}
\end{figure}

In this work, we propose \name, an approach that bridges this gap by treating classification as a constrained generation task with \textbf{single-token outputs}. More broadly, we view classification as one member of a family of \emph{atomic actions}---fast, ``System 1'' decisions that software can consume directly, such as classification, semantic endpointing, Boolean checks, and scoring---all of which reduce to producing a single typed token in one step. Our key insight is to introduce atomic special tokens (e.g., \texttt{[control\_1]}) for each class, enabling the model to produce decisions in exactly one generation step. As illustrated in Figure~\ref{fig:main_idea}, this design not only eliminates multi-token decoding overhead but also enables \emph{zero-shot classification}: by reassigning label tokens at inference time, the model seamlessly adapts to new tasks without task-specific retraining.

We demonstrate this approach using supervised fine-tuning on large language models, creating a unified framework that handles both text-only and multimodal classification tasks. Our method rests on three central pillars: \textbf{accuracy}, \textbf{latency}, and \textbf{generality}. By constraining the output space to a finite set of learned tokens, we achieve deterministic single-step inference while maintaining the semantic understanding capabilities of large pretrained models. Our contributions are threefold:

\begin{enumerate}
\item \textbf{A unified single-token classification framework} that integrates single-step constrained generation, randomized label assignment, and multimodal inputs into one deployable interface, removing multi-token decoding overhead while retaining the broad generalization of decoder LLMs.

\item \textbf{Frontier-level accuracy at a fraction of the latency.} Under an identical, directly comparable serving setup, \name{} matches the accuracy of the strongest frontier LLMs (Claude-4.8-Opus, Gemini-Pro-3.1) while answering in sub-second time---where those frontier models take one to four seconds at the median---with a near-flat tail that holds steady as the label space grows.

\item \textbf{Validation across real deployment settings, alongside comprehensive public benchmarks.} Beyond standard public datasets (text-only and multimodal), we evaluate \name on a real production workload from a conversational agent---a de-identified intent-routing task whose label inventory evolves over time---and on a latency-critical synthetic voice turn-taking / endpointing task where sub-second tail latency is a hard requirement, showing the framework holds under real production constraints, not just academic ones.
\end{enumerate}

\section{Related Work}

\subsection{Prompt-Based Classification with LLMs}
LLMs are often adapted for classification via prompting and in-context learning (further strengthened by instruction tuning)~\citep{brown2020language,sun2023text, ouyang2022training,wei2021finetuned}, yet their outputs can remain free-form and multi-token, complicating structured integration and increasing inference cost.

To mitigate this, methods such as PET and LM-BFF reformulate classification as a cloze task with \emph{verbalizers} that map each class to a natural-language string and fine-tune the model on small labeled sets~\citep{schick2020exploiting, gao2020making}.  Beyond discrete verbalizers, 
\emph{continuous prompting} methods have been proposed: Prompt Tuning learns
soft prompt embeddings optimized for a target task
\citep{lester2021prompt,ding2023parameter}, while Prefix Tuning prepends trainable key–value 
vectors to each transformer layer \citep{li2021prefix}. Although effective, these approaches are typically developed for few-shot learning scenarios with limited labeled data; in contrast, our work targets settings with more training examples.

Recent work uses \emph{constrained decoding} to restrict generations to valid label strings~\citep{geng2023grammar}, and a mature ecosystem of structured-generation engines now enforces such constraints efficiently at the inference layer~\citep{willard2023outlines,dong2024xgrammar,zheng2023sglang}. However, these methods constrain \emph{which} tokens may be emitted while still decoding label strings token by token, so their cost grows with label-string length and can be inefficient for large label spaces or multimodal settings; imposing output formats can also degrade task performance~\citep{tam2024letmespeak}. We instead use a finite set of atomic label tokens, so that a valid decision is guaranteed in a single decoding step regardless of how the label reads as text.


\subsection{Encoder Fine-Tuning for Classification}
A widely adopted paradigm emphasizes \emph{fine-tuning encoder models} with a
task-specific classifier head. Transformer encoders such as BERT and its successors
(e.g., RoBERTa, DeBERTa) have become the standard for text classification~\citep{devlin2019bert,liu2019roberta,he2020deberta}.
For multimodal classification, encoder-style fusion models extend this paradigm by incorporating vision or audio encoders and cross-modal attention modules, as in MulT and MAG-BERT~\citep{tsai2019multimodal,rahman2020integrating}.
These approaches are efficient and effective, but they require task-specific
classifier heads and lack the flexibility of LLMs.
In contrast, \name shows that \emph{decoder-style VLMs}, adapted with single-token label spaces, can match or surpass encoder baselines on multimodal datasets while retaining the generalization of decoder LLMs, and offers a \emph{systematic framework} validated in a real production deployment. A complementary line of work instead repurposes decoder LLMs \emph{as} encoders for classification and embedding, either by extracting last-token representations for a lightweight head or by unlocking bidirectional attention~\citep{behnamghader2024llm2vec,yousefiramandi2025finetuning}; these methods still attach a separate task-specific head, whereas \name keeps a single generative interface and emits the decision as one output token.

\subsection{Single-Token and Restricted-Vocabulary Classification}
A recent line of work---concurrent with and independent of ours---reformulates classification so that a decision is read out in a single forward pass rather than generated token by token. Most closely related, SALSA~\citep{berdichevsky2025salsa} restricts the output distribution to per-class tokens and adapts the model with parameter-efficient fine-tuning, reporting strong accuracy at low cost. Our framework differs in two respects that matter for deployment: we \emph{randomize the label-to-token assignment during training}, decoupling each control token from fixed class semantics and enabling zero-shot adaptation to new or relabeled class sets at inference; and we treat \emph{latency as a first-class object of study}, reporting full P50/P95 tail-latency under an identical, directly comparable serving setup against a broad panel of frontier LLMs, which prior single-pass classifiers omit.

This single-token, typed-decision paradigm has also attracted industry attention. Commercial systems such as TypeSafe AI's ``System One'' models target instant, low-latency decisions that software can consume directly~\citep{typesafe2026systemone}, and independent reproductions report calibrated one-pass readouts from small LoRA-tuned backbones~\citep{ren2026openjev}. These efforts underscore the practical demand for the problem we study, but they differ from \name in scope and evidence: the commercial training recipe is proprietary, and---most relevant to our setting---they address \emph{single-label, text-only} typed decisions and do not target multimodal inputs or multi-label outputs. The approach we describe was publicly disclosed and deployed earlier (\href{https://openreview.net/forum?id=78WdKlYSeO}{OpenReview}, September 2025; \href{https://engineering.salesforce.com/solving-real-time-ai-classification-for-agentforce-how-single-token-prediction-delivers-30x-faster-agent-responses/}{Salesforce Engineering}, November 2025); this paper formalizes it, extends it to multimodal and multi-label classification, and characterizes its latency behavior against strong baselines under real production constraints.

\section{Methodology}

\subsection{Problem Formulation}

We formalize classification as a supervised learning problem. 
Each training instance consists of an input tuple
\[
x = \bigl(x^{\text{text}}, \, x^{\text{vis}}\bigr),
\]
where $x^{\text{text}}$ denotes the textual input and 
$x^{\text{vis}}$ represents the vision modality, which may include static images 
or short video segments. Not all instances contain both modalities; missing 
modalities are treated as empty. We aim to learn a mapping
\[
f_\theta : \bigl(x^{\text{text}}, \, x^{\text{vis}}\bigr) \;\mapsto\; y,
\]
where $y \in \mathcal{Y}$ is a categorical label drawn from a predefined set of 
classes $\mathcal{Y} = \{c_1, c_2, \dots, c_K\}$. 

Given a dataset $\mathcal{D} = \{(x_i, y_i)\}_{i=1}^N$,
the learning goal is to estimate model parameters $\theta$ that minimize the 
expected classification loss:
\[
\theta^* = \arg\min_{\theta} \; \mathbb{E}_{(x,y)\sim \mathcal{D}} 
\;\mathcal{L}\bigl(f_\theta(x), \, y\bigr),
\]
where $\mathcal{L}$ is the cross-entropy loss over labels. This formulation generalizes unimodal classification to a multimodal setting, where textual and visual inputs are jointly leveraged for improved prediction.

\subsection{Single-Token Classification with Atomic Label Tokens}

Our proposed framework \name is model-agnostic and can be applied to a wide range of LLMs and VLMs, including the Gemma, Mistral, and GPT-OSS architectures we use here, among others. The general principle is to treat classification as a constrained generation task, where the model is guided to produce a single special token corresponding to the target label.


\paragraph{Special Tokens for Labels.}
For each classification category $c_k \in \mathcal{Y}$, we introduce a unique 
special token $[o_k]$. Conventional choices such as digits or short text labels are problematic: they can 
appear naturally in the input, causing ambiguity, and larger indices (e.g., ``100'') are split into multiple subword tokens, leading to multi-step decoding and higher latency. To provide grounding, a natural-language description of each class is included in the system prompt (see Appendix~\ref{app:prompt_configuration} for prompt templates). During fine-tuning, the model is trained to map $\bigl(x^{\text{text}}, x^{\text{vis}}\bigr)$ directly to the correct special token.



Let $\mathcal{Y} = \{c_1, \ldots, c_K\}$ denote the label set and let $\mathcal{V}$ be the
tokenizer vocabulary. We augment the vocabulary with $K$ \emph{atomic output symbols}:
\[
\Omega = \{[o_1], [o_2], \ldots, [o_K]\}, 
\qquad 
\tilde{\mathcal{V}} = \mathcal{V} \cup \Omega.
\]
Each $[o_k]$ is a dedicated \emph{control token} corresponding to class $c_k$, with a trainable
embedding $e_k \in \mathbb{R}^d$. These rows are appended to the model’s embedding matrix and are updated during fine-tuning, together with the adapted model weights. In implementation, we pre-allocate a pool of up to 500 reserved control tokens, allowing the framework to support classification tasks with as many as 500 labels without retraining.

\paragraph{Why single-token outputs.}
If labels are represented as natural-language strings $s_k$ (e.g., \texttt{"book a flight"}), a
subword tokenizer typically produces a variable-length sequence
\[
\tau(s_k) = (v_1, \ldots, v_{m_k}), \quad m_k \geq 1,
\]
where $\tau(\cdot)$ denotes the tokenization function. This variability requires loss over multiple
decoding steps and depends on segmentation. In contrast, our design assigns each class a single
atomic symbol $\omega_k = [o_k] \in \Omega$, which is guaranteed to decode as exactly one token.
This yields three benefits:  
(i) the output space collapses to $K$ symbols,  
(ii) ambiguity from label strings that may appear in the input is eliminated, and  
(iii) decoding and evaluation are simplified to a deterministic one-step classification.

\paragraph{Randomized label assignments.}
To prevent memorization of static token--label associations, we do not fix a permanent mapping 
between special tokens and semantic classes. Instead, during preprocessing we randomly shuffle 
the correspondence between classes and control tokens across training instances. This design encourages the model to rely on the contextual descriptions of labels provided in the prompt, rather than memorizing token identities. As a result, the model learns to infer the correct output token from the input context, which improves robustness and generalization across datasets and label spaces. Without randomization, models tended to overfit to token IDs and failed to generalize to new mappings.

\paragraph{Separation from context.}
We reserve a sentinel namespace for $\Omega$, ensuring that these tokens are never decomposed
into subwords and never occur in the input text. Formally, $\Omega \cap \mathcal{V} = \emptyset$
and $\Omega \cap \tau(x) = \emptyset$ for any input $x$. In practice, this is enforced by
registering $[o_k]$ as special tokens in the tokenizer,
so that they are available to the model during output generation.

\paragraph{Training objective (single position).}
Given input $x = \bigl(x^{\text{text}}, \, x^{\text{vis}}\bigr)$ with gold label $k \in \{1, \ldots, K\}$,
the model is required to emit exactly one control token $[o_k]$. Let $h_T$ denote the decoder
state at the output position (the single assistant step). We compute logits restricted to $\Omega$:
\[
\begin{aligned}
z_j &= (W h_T + b)_j, \\
P_\theta([o_j] \mid x)
&= \frac{\exp(z_j)}{\sum_{i=1}^K \exp(z_i)},
\quad j \in \{1, \ldots, K\}.
\end{aligned}
\]

The loss is standard cross-entropy over this single prediction:
\[
\mathcal{L}(x,k) = - \log P_\theta([o_k] \mid x).
\]
All preceding tokens (system/user prompts and any in-context descriptions) are masked out
of the loss, so classification supervision is concentrated solely on the final output position.
In implementation, we construct a binary loss mask: the position of the response token is assigned its class label,
while all other positions are set to \(-100\), which the cross-entropy loss ignores.

\paragraph{Inference rule.}
At test time, decoding reduces to a single restricted argmax:
\[
k^* = \arg\max_{k \in \{1, \ldots, K\}} P_\theta([o_k] \mid x).
\]
Equivalently, we set \texttt{max\_new\_tokens = 1} and restrict the decision space to $\Omega$,
ensuring the model outputs exactly one control token (e.g., $[o_{k^*}]$) rather than a multi-token string.
This deterministic procedure avoids token-by-token generation: the \emph{decode} stage is a single forward pass whose cost is independent of output length, unlike autoregressive decoding. The \emph{prefill} stage still scales with input and prompt length, so our $\mathcal{O}(1)$ claim refers specifically to decoding.

\subsection{Extension to Multi-Label Classification}
\label{sec:multilabel}
In the settings considered so far, each input is assigned exactly one class ($y \in \mathcal{Y}$). Many real-world applications are instead \emph{multi-label}: a single input may belong to several classes at once, so the prediction target is a \emph{subset} $Y \subseteq \mathcal{Y}$ rather than a single label. For instance, a customer utterance may simultaneously request an action, reference a prior case, and fall under a compliance category. Autoregressive approaches handle this awkwardly, because generating labels one at a time is known to suppress all but one label at each step, a bias that supervised fine-tuning tends to amplify~\citep{ma2025multilabel}. Our single-token formulation sidesteps this pathology by construction: the restricted softmax over $\Omega$ defined above already yields, in a single forward pass, a full probability distribution over the entire label-token set, so the relative evidence for \emph{all} candidate labels is available simultaneously, without any sequential generation.

The key enabler is a change in \emph{training supervision} rather than in architecture or decoding. Instead of supervising the restricted distribution toward a one-hot target concentrated on a single control token, we train the model to match a \emph{soft target distribution} that spreads probability mass across the control tokens of all labels relevant to an instance. Concretely, the model minimizes a distribution-matching objective---a divergence between the predicted restricted distribution over $\Omega$ and this target---instead of the single-label cross-entropy used above. This teaches the model to allocate calibrated probability mass over co-occurring labels while preserving the single-pass, single-step decoding cost: multi-label prediction requires no additional generation steps and no architectural changes, only a different training target. At inference, the same one-pass distribution is used to select the set of labels the model assigns sufficient probability to, rather than a single \mbox{arg\,max}. We omit the precise construction of the target distribution and the selection rule; our aim here is to show that the single-token framework \emph{extends naturally} to multi-label settings, inheriting its latency and determinism guarantees.

\section{Experiments}

\subsection{Datasets}
We adapt each base model into a single-token classifier with supervised fine-tuning on a corpus of classification instances covering both text-only and multimodal inputs, and then evaluate the adapted models on the held-out benchmarks below; all reported evaluation datasets are disjoint from any data used for adaptation.

\paragraph{Evaluation benchmarks.}
We evaluate our approach on a diverse suite of multimodal and text-only classification benchmarks. For multimodal evaluation, we use the official test split of \textbf{MIntRec2.0}~\citep{zhang2024mintrec2}, a large-scale benchmark for intent recognition in multimodal dialogues that combine text and vision. This dataset includes 30 fine‐grained intent classes and requires reasoning over conversational context and multiple modalities, making it a challenging test of real-world multimodal classification. For text-only tasks, we evaluate on two widely used public sentiment benchmarks: \textbf{SST-2}~\citep{socher2013recursive}, a binary sentiment analysis benchmark of movie reviews annotated as positive or negative, and \textbf{Amazon Reviews Polarity}~\citep{mcauley2013hidden, zhang2015character}, which contains millions of product reviews labeled as positive or negative and probes sentiment classification in a large-scale, noisy e-commerce domain. We evaluate on the \emph{full} SST-2 test set; for Amazon Reviews Polarity, whose test set is much larger, we evaluate on 5k examples of the official test split. We further include an \textbf{intent routing} dataset derived from a production conversational agent, where each example is labeled with an intent that triggers a predefined routing action (mapping the request to a follow-up or tool call); we report only aggregate results.\footnote{We cannot release this dataset due to privacy and contractual restrictions; all reported numbers are aggregated.} Finally, to go beyond sentiment into a modern deployment setting, we add a \textbf{voice turn-taking / endpointing} benchmark (353 examples, binary): given a partial conversation transcript, the model must decide whether the speaker has finished their turn (so the agent should respond) or is still speaking (so the agent should keep listening). We further report results on additional public benchmarks (DBpedia~\citep{lehmann2015dbpedia}, TweetTopic, Banking77, and MTOP) in Appendix~\ref{app:additional_benchmarks:other}. Example prompt templates are provided in Appendix~\ref{app:prompt_configuration} (\Cref{fig:dataset_example_positive,fig:dataset_example_amazon,fig:dataset_example_dbpedia,fig:dataset_example_topic_thank}).

\subsection{Baselines}
We compare against both open-source models and commercial API systems. \textbf{Pretrained LLMs.} We adapt open-weight base models from the Gemma, Mistral, and GPT-OSS families, and also evaluate their un-adapted checkpoints to quantify the performance of large pretrained models without task-specific adaptation (exact checkpoints are given with each results table). \textbf{Commercial API models.} As strong performance benchmarks, we evaluate a broad panel of frontier LLMs spanning the major providers: GPT-4o, GPT-4.1, GPT-4.1-mini, and GPT-5 (OpenAI); Claude-4-Sonnet, Claude-5-Sonnet, and Claude-4.8-Opus (Anthropic); and Gemini-2.5-Pro and Gemini-Pro-3.1 (Google). 
\textbf{Dedicated single-token systems.} We further compare against Jev/TypeSafe, a commercial system purpose-built for single-token typed decisions, on the tasks where its interface applies. 
\textbf{Encoder-based models.} To contextualize against specialized architectures, we also report results for strong encoder-style multimodal baselines (e.g., \texttt{MAG-BERT} and \texttt{MulT}), following the \textsc{MIntRec}~2.0 benchmark protocol. For text benchmarks, we include: (i) a traditional linear classification head applied to BERT-base and RoBERTa-base (16-shot fine-tuning), and (ii) LM-BFF~\citep{gao2020making} with a RoBERTa-base backbone as a representative prompt-based few-shot classifier.

\subsection{Evaluation metrics}
We evaluate models along two complementary dimensions. \textbf{Classification accuracy.} The primary metric is accuracy, computed as the percentage of instances where the predicted control token matches the gold label. Accuracy is reported separately for each dataset. \textbf{Latency.}
To capture serving efficiency, we report the median (P50) and tail (P95) latency of a single request. Comparing systems as different as our single-token classifier and commercial frontier LLMs requires care, so we access every model through the \emph{same} serving interface and issue requests one at a time in round-robin order: each model faces identical load and its measured end-to-end time is directly comparable to the others, with no batching and no client-side caching (\S\ref{sec:latency}). We separately examine how latency scales with batch size using our own vLLM-based inference framework~\citep{kwon2023efficient}; Appendix~\ref{app:sensitivity_analysis:batch} presents a batch-scaling study demonstrating that the latency benefits of \name\ persist even under large-batch inference settings.

\subsection{Results}

\subsubsection{Multimodal evaluation results}
\label{sec:results_multimodal}

We evaluate our fine-tuned models trained on the challenging MIntRec~2.0 dataset, which requires understanding multimodal dialogue contexts (text, image, video) for intent recognition. 

\paragraph{MIntRec~2.0 (Multimodal Topic Classification).}
Table~\ref{tab:mintrec_results} presents results on MIntRec~2.0. Evaluated zero-shot under a natural label-name prompt, base VLMs reach only $\sim$35--43\% accuracy, underscoring the difficulty of multimodal intent recognition without adaptation. After fine-tuning, performance improves markedly: Gemma-3-4B reaches \textbf{55.2\%}, while Gemma-3-27B achieves \textbf{62.7\%}. Crucially, this margin holds not only over the original GPT-4o (42.1\%) and GPT-5 (48.5\%) baselines but against the strongest frontier VLMs evaluated under the identical prompt \emph{with} the same video frames: Gemini-Pro-3.1 (55.9\%), Claude-4.8-Opus (52.9\%), and Claude-5-Sonnet (49.8\%) all remain below our fine-tuned Gemma-3-27B, and even our 4B \name{} model (55.2\%) surpasses every frontier VLM except Gemini-Pro-3.1. Even with the video frames available, frontier VLMs plateau well below our adapted models, consistent with our motivation that the visual signal needs task-specific multimodal adaptation to become usable.

Beyond this standard GPT-4o configuration, we also explore whether carefully designed class-index prompts can emulate our single-token interface; Appendix~\ref{app:preliminary_experiment:alt_prompt} (\emph{Alternative Prompting Baselines}) shows that such prompting-only variants still frequently produce multi-token outputs and achieve only 45.12\% accuracy. To further strengthen this comparison, we additionally evaluate a constrained-decoding variant of GPT-4o, where all 30 intents are mapped to digit-only labels and decoding is restricted to this label set. This improves GPT-4o's accuracy to 45.56\%, but it still falls well short of \name-adapted open-source models.

\paragraph{Comparison with Encoder-Based Models.}
MAG-BERT and MulT, strong encoder-based multimodal baselines, achieve around 60.6\% accuracy. Our fine-tuned Gemma-3-27B surpasses both, reaching \textbf{62.7\%}, while also offering the benefits of a unified generalization modeling framework. However, these encoder-based models are fully fine-tuned specifically on MIntRec~2.0 with task-specific classifier heads tied to a fixed label space, and therefore do not generalize to new tasks or label sets without retraining. In contrast, \name models preserve the generalization interface of decoder LLMs and support zero-shot adaptation across tasks; we further analyze this generality in Section~\ref{sec:text_cls}. The competitive or superior performance of fine-tuned LLMs indicates that large decoder-based architectures, when fine-tuned on in-domain data, can match or outperform specialized encoder-based models while simultaneously enabling broader generalization and reasoning capabilities.

\begin{table}[t]
\centering
\small
\setlength{\tabcolsep}{5pt}
\renewcommand{\arraystretch}{1.05}
\begin{tabular}{lc}
\toprule
\textbf{Model} & \textbf{Acc.} \\
\midrule
\multicolumn{2}{l}{\emph{Base VLMs (no adaptation, zero-shot)}} \\
Mistral-3-24B (Base) & 35.17 \\
Gemma-3-4B (Base)      & 36.65 \\
Gemma-3-27B (Base)     & 43.29 \\
\midrule
\multicolumn{2}{l}{\emph{Frontier VLMs (zero-shot, +video)}} \\
GPT-4o                 & 42.06 \\
GPT-4.1                & 45.40 \\
GPT-4.1-mini           & 45.99 \\
GPT-5                  & 48.45 \\
Claude-4-Sonnet        & 48.55 \\
Gemini-2.5-Pro         & 49.14 \\
Claude-5-Sonnet        & 49.83 \\
Claude-4.8-Opus        & 52.88 \\
Gemini-Pro-3.1         & 55.93 \\
\midrule
\multicolumn{2}{l}{\emph{Encoder baselines (task-specific FT)}} \\
MAG-BERT               & 60.58 \\
MulT                   & 60.66 \\
\midrule
\multicolumn{2}{l}{\emph{Fine-tuned (\name, ours)}} \\
Mistral-3-24B (FT, \name) & 49.34 \\
Gemma-3-4B (FT, \name)    & 55.19 \\
\rowcolor{gray!15}
\textbf{Gemma-3-27B (FT, \name)} & \textbf{62.72} \\
\bottomrule
\end{tabular}
\caption{MIntRec~2.0 accuracy (\%), all models given text + video. Base VLMs and
frontier VLMs are evaluated zero-shot under the \emph{same} natural-language prompt
(list the 30 intent names, reply with one). The strongest zero-shot model
(Gemini-Pro-3.1, 55.9\%) still trails our fine-tuned Gemma-3-27B (62.7\%); even our
4B \name{} model (55.2\%) surpasses every frontier VLM except Gemini-Pro-3.1.}
\label{tab:mintrec_results}
\end{table}

\subsubsection{Text classification across domains}
\label{sec:text_cls}

\begin{table*}[t]
\centering
\small
\setlength{\tabcolsep}{6pt}
\renewcommand{\arraystretch}{1.1}
\begin{tabular}{lcccc}
\toprule
\textbf{Model} & \textbf{SST-2} & \textbf{Amazon} & \textbf{Intent Routing} & \textbf{Voice} \\
\midrule
\multicolumn{5}{l}{\emph{Encoder / few-shot baselines (task-specific)}} \\
BERT (linear head)     & 70.87 & 77.04 & --    & --    \\
RoBERTa (linear head)  & 61.35 & 79.16 & --    & --    \\
LM-BFF                 & 90.14 & 83.76 & --    & --    \\
\midrule
\multicolumn{5}{l}{\emph{Ours --- \name{} single-token classifiers (zero-shot on every set)}} \\
\rowcolor{gray!15}
Gemma-4-26B (FT, \name)    & 95.76 & 96.60 & 81.43 & 94.33 \\
\rowcolor{gray!15}
GPT-OSS-20B (FT, \name)    & 92.89 & 95.34 & 85.52 & 96.90 \\
\midrule
\multicolumn{5}{l}{\emph{Frontier LLMs (zero-shot)}} \\
GPT-4o                 & 95.64 & 96.32 & 84.84 & 95.75 \\
GPT-4.1                & 95.99 & 96.56 & 86.03 & 96.03 \\
GPT-4.1-mini           & 94.61 & 96.28 & 69.34 & 94.62 \\
GPT-5                  & 95.64 & 96.40 & 83.99 & 96.88 \\
Claude-4-Sonnet        & 95.41 & 96.78 & 86.03 & \textbf{97.17} \\
Claude-5-Sonnet        & 95.76 & 97.28 & 85.69 & 96.88 \\
Claude-4.8-Opus        & 96.33 & 97.14 & \textbf{87.73} & 96.60 \\
Gemini-2.5-Pro         & 95.64 & 96.68 & 84.50 & 96.60 \\
Gemini-Pro-3.1         & \textbf{96.79} & \textbf{97.42} & 87.05 & 95.18 \\
\midrule
\multicolumn{5}{l}{\emph{External evaluator (public + synthetic only)}} \\
Jev (TypeSafe)         & 96.10 & 96.80 & --    & 93.20 \\
\bottomrule
\end{tabular}
\caption{Text classification accuracy (\%) on SST-2 (872 examples), Amazon Reviews (5k),
a production intent-routing benchmark (587 utterances, with a label
inventory that evolves over time), and a voice turn-taking / endpointing benchmark (353 examples, binary).
\textbf{Every LLM row---ours and the frontier models---is evaluated zero-shot; no model in
this table was trained on any of these datasets.} Our \name{} single-token classifiers sit
inside the frontier band on every set (within $\sim$1--2 points of the best model) and
lead the smaller/deployable tier, at a fraction of the serving latency
(Figure~\ref{fig:text_latency_p50}). Encoder / few-shot baselines are task-specific and are
not applicable to the intent-routing benchmark (its label space evolves over time, requiring
retraining) or to voice endpointing. Jev is not run
on the intent-routing set (internal data). \textbf{Bold} = best per column.}
\label{tab:textonly-multidataset}
\end{table*}

We evaluate on two widely used public sentiment benchmarks (SST-2, Amazon Reviews), an intent-routing dataset, and a voice turn-taking / endpointing benchmark. Evaluations on additional public benchmarks (DBpedia, TweetTopic, Banking77, and MTOP~\citep{li2021mtop}) are provided in the appendix (Appendix~\ref{app:additional_benchmarks:other} and the scaling/cardinality analyses).
We assess cross-domain transfer by evaluating our \emph{pre-adapted} single-token classifiers zero-shot on these datasets, and compare them against a broad panel of frontier LLMs (also zero-shot), encoder/few-shot baselines, and an external commercial evaluator. Table~\ref{tab:textonly-multidataset} reports the per-dataset accuracy results, and Figure~\ref{fig:text_latency_p50} summarizes end-to-end serving latency across the same models.

Notably, the intent-routing setting differs from public benchmarks in that the topic inventory evolves over time (e.g., new intents are introduced and old ones are merged or deprecated).
This makes classifier-head baselines and dataset-specific verbalizers tied to a fixed label space and thus costly to maintain, whereas \name's reserved atomic label tokens support one-step routing with prompt-side label updates. On this intent-routing benchmark, \name reaches \textbf{85.5\%} accuracy, competitive with the frontier panel (within $\sim$2 points of the strongest, Claude-4.8-Opus and Gemini-Pro-3.1), while serving each request in 0.53\,s at the median (p95 0.63\,s)---$\sim$1.9--7.5$\times$ faster at the median (and $\sim$2.4--9.0$\times$ at the P95 tail) than these frontier models under identical serving conditions (Figure~\ref{fig:text_latency_p50}). We next evaluate on public benchmarks with fixed label spaces.

As a basic encoder-style baseline, we include BERT-base and RoBERTa-base models equipped
with a linear classification head, fine-tuned using 16 examples per class. As shown in Table~\ref{tab:textonly-multidataset}, these linear-head models exhibit the weakest accuracy, and---being tied to a fixed label space---do not transfer to the intent-routing or voice-endpointing settings.
To provide a stronger encoder-based comparison, LM-BFF augments RoBERTa with
prompt-based verbalizers and in-context examples. LM-BFF clearly improves over linear-head
fine-tuning, reaching 83.8--90.1\% accuracy on SST-2 and Amazon Reviews. However, it still falls short of our \name models, which reach 95.8\% on SST-2 and 96.6\% on Amazon Reviews.
Moreover, LM-BFF relies on multi-token prompted decoding and, like the linear-head baselines, requires task-specific fine-tuning for each dataset, so it does not transfer to new or evolving label sets.

Without task-specific adaptation, \name demonstrates \textbf{robust cross-domain
performance}. As shown in Table~\ref{tab:textonly-multidataset}, our fine-tuned models are \textbf{competitive with the frontier panel} on sentiment classification (SST-2, Amazon Reviews) and stay within $\sim$1--2 points of the strongest frontier models on every set, while leading the smaller/deployable tier on the intent-routing benchmark. They match this accuracy at a small fraction of the serving latency---the subject of the next section (\S\ref{sec:latency}). Additional analyses on label-set cardinality and model scaling are deferred to Appendix~\ref{app:sensitivity_analysis:label} and Appendix~\ref{app:sensitivity_analysis:scale}.

We further examine whether the model operates under unseen label--token mappings (Appendix~\ref{app:sensitivity_analysis:zero-shot}); a consolidated accuracy--latency comparison is given in Figure~\ref{fig:pareto_enterprise}.

\subsubsection{Serving latency}
\label{sec:latency}
\begin{figure}[t]
    \centering
    \includegraphics[width=\columnwidth]{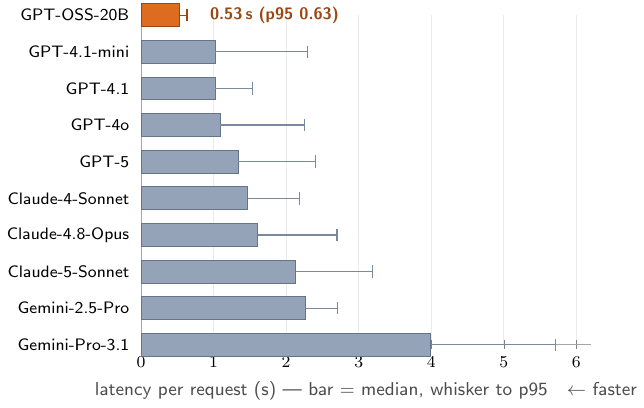}
    \caption{End-to-end per-request serving latency on the intent-routing
    set (587 utterances), median (p50) with a whisker to the p95 tail.
    Orange = our \name{} single-token classifier; slate = frontier LLMs. Every model is
    accessed through the same serving interface and issued the same requests one at a time
    in round-robin order, so all models see identical load and their times are directly
    comparable. Our classifier answers in
    \textbf{0.53\,s} (p50) --- $\sim$1.9$\times$ faster than the quickest frontier model
    (GPT-4.1-mini) and $\sim$7.5$\times$ faster than the slowest (Gemini-Pro-3.1) --- and
    has a near-flat tail (p95~$=$~0.63\,s) where frontier models fan out by 0.5--1.7\,s.}
    \label{fig:text_latency_p50}
\end{figure}
Accuracy places \name{} inside the frontier band; latency is where the single-token
formulation separates from it. To make the comparison fair, we access every model---\name{}
and all frontier LLMs---through the \emph{same} serving interface and issue requests one at a
time in round-robin order, so each model faces identical load and its measured time is
directly comparable to the others; we report the honest end-to-end time of a single request
(no batching, no client-side caching). Because \name{} emits exactly one token, its
end-to-end time is essentially its time-to-first-token; frontier models instead pay for
multi-token generation, and that difference is the entire point.

\paragraph{Sub-second serving with a flat tail.} On the intent-routing set
(Figure~\ref{fig:text_latency_p50}), \name{} answers in \textbf{0.53\,s} at the median and
\textbf{0.63\,s} at the P95---from $\sim$1.9$\times$ faster than the quickest frontier model
(GPT-4.1-mini) to $\sim$7.5$\times$ faster than the slowest (Gemini-Pro-3.1). Just as important as the median is the
\emph{shape} of the distribution: \name's tail is nearly flat (P95$-$P50\,$\approx$\,0.10\,s),
whereas frontier models fan out by 0.5--1.7\,s from median to P95, so the speed gap only widens at the tail. Under real load and
retries it is the tail, not the median, that governs user-visible latency, so this gap widens
in deployment.

\paragraph{Latency is independent of the task.} Because a decision is always one forward step,
\name's latency barely moves across tasks of very different label-space size and content: its
median is 0.529\,s on the evolving intent router, 0.526\,s on binary SST-2 sentiment, and
0.527\,s on voice turn-taking / endpointing (Figure~\ref{fig:voice_latency};
SST-2 in Appendix~\ref{app:additional_benchmarks:latency}, Figure~\ref{fig:sst2_latency})---the three medians fall within
3\,ms of one another despite spanning a large evolving label set, binary sentiment, and a
real-time transcript decision. Autoregressive baselines have no such guarantee---their cost
scales with how many tokens they choose to emit, which is why their tails are both higher and
more variable. This predictability, not just the average speed, is what makes the approach
suitable for latency-critical routing.

\paragraph{The speedup is intrinsic to single-token decoding.} A useful control is the
external commercial single-token system Jev/TypeSafe, which on the voice endpointing task
serves at 0.55\,s median (P95 0.69\,s), essentially tied with \name{}
(Figure~\ref{fig:voice_latency})---and well ahead of every frontier LLM, which must decode
over the running dialogue transcript and lands at 0.98--2.88\,s at the median. This is the
expected result and reinforces our thesis: the several-fold latency advantage comes from
collapsing classification into a single typed decision, not from any implementation trick
particular to our system. Unlike Jev, however, \name{} delivers this while also handling
multimodal inputs and multi-label outputs.

\begin{figure}[t]
    \centering
    \includegraphics[width=\columnwidth]{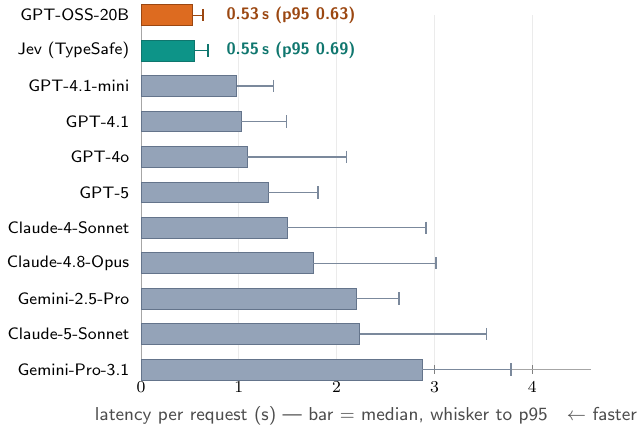}
    \caption{End-to-end per-request serving latency on the voice turn-taking / endpointing
    benchmark (353 examples), under the same round-robin, concurrency-1, shared
    serving-interface protocol as Figure~\ref{fig:text_latency_p50}.
    Orange = our \name{} single-token classifier; teal = the external single-token system
    Jev/TypeSafe; slate = frontier LLMs. \name{} answers in 0.53\,s (p95 0.63\,s), matching its
    latency on the intent-routing and SST-2 tasks almost to the millisecond, and the external
    single-token system Jev/TypeSafe is essentially tied (0.55\,s, p95 0.69\,s), while every
    frontier model must decode over a running dialogue transcript and lands at 0.98--2.88\,s at
    the median.}
    \label{fig:voice_latency}
\end{figure}

\begin{figure}[t]
    \centering
    \includegraphics[width=\columnwidth]{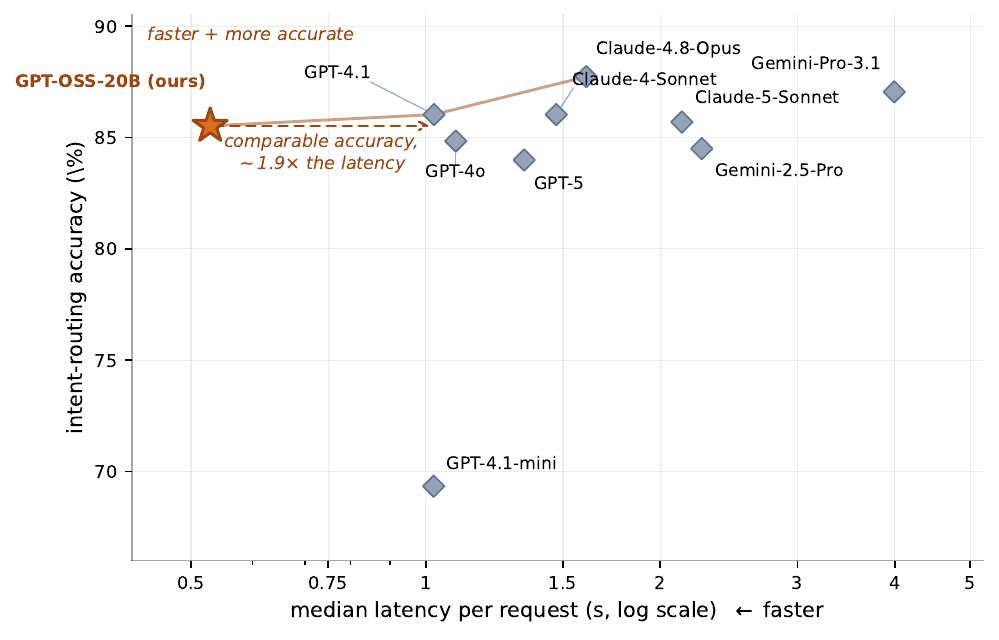}
    \caption{Accuracy vs.\ median serving latency on the intent-routing benchmark (log-scaled latency; lower-left of the accuracy axis omitted). All points use the same round-robin serving protocol. Our single-token classifier (orange star) occupies the fast, high-accuracy corner; the orange curve traces the Pareto frontier (\name{}, GPT-4.1, Claude-4.8-Opus), and the dashed arrow marks the nearest frontier point of comparable accuracy (GPT-4.1), which runs at $\sim$1.9$\times$ our latency. No model is both faster and more accurate than \name{}.}
    \label{fig:pareto_enterprise}
\end{figure}

\paragraph{Accuracy and latency together.} Figure~\ref{fig:pareto_enterprise} unifies the two
axes on the intent-routing benchmark, with every model measured under the identical round-robin
protocol (single serving interface, one request at a time, end-to-end wall-clock). Our
single-token classifier sits alone in the fast corner at $0.53$\,s and $85.5\%$ accuracy, and
anchors the fast end of the Pareto frontier, which it shares with just two other models:
GPT-4.1, which edges ahead on accuracy ($86.0\%$) at $\sim$1.9$\times$ our latency, and
Claude-4.8-Opus, the most accurate ($87.7\%$) at $\sim$3$\times$. A few more models are also more
accurate but lie off the frontier because a frontier model already dominates them---Claude-4-Sonnet
($86.0\%$), Claude-5-Sonnet ($85.7\%$), and even Gemini-Pro-3.1 ($87.1\%$), which pays $\sim$7.5$\times$
our latency yet is still edged out on accuracy by Claude-4.8-Opus. The remaining frontier LLMs are
strictly dominated by \name{} itself---slower without being more accurate (GPT-4o, GPT-5, and
Gemini-2.5-Pro all trail on both axes). No model is both faster and more accurate than \name{},
which places our system at the knee of the frontier rather than at either extreme.

\section{Conclusion}
We introduced \name, a framework for structured decision making that casts classification as single-step constrained generation using atomic label tokens and supervised fine-tuning. Under an identical, directly comparable serving setup, \name{} matches the accuracy of the strongest frontier LLMs across sentiment, intent routing, and multimodal intent recognition, yet answers in sub-second time with a near-flat tail---and, because every decision is a single decode step, its latency stays essentially constant (0.53\,s at the median) across tasks whose label spaces and inputs differ sharply. This combination of frontier-level accuracy, several-fold lower latency, and task-independent predictability is what makes the framework practical under real production constraints; extending beyond text and vision and improving multilingual generalization are key directions.


\section*{Ethical Considerations}

The intent-routing evaluation uses de-identified interactions from a production conversational agent. We report only aggregate metrics, include no raw utterances, and cannot release the data due to privacy and contractual restrictions. The intended use is intent routing, not user profiling or high-stakes automated decisions.

\bibliography{custom}

\begin{thebibliography}{40}
\providecommand{\natexlab}[1]{#1}

\bibitem[{Antypas et~al.(2022)Antypas, Ushio, Camacho-Collados, Silva, Neves, and Barbieri}]{antypas2022twitter}
Dimosthenis Antypas, Asahi Ushio, Jose Camacho-Collados, Vitor Silva, Leonardo Neves, and Francesco Barbieri. 2022.
\newblock Twitter topic classification.
\newblock In \emph{Proceedings of the 29th International Conference on Computational Linguistics}, pages 3386--3400.

\bibitem[{BehnamGhader et~al.(2024)BehnamGhader, Adlakha, Mosbach, Bahdanau, Chapados, and Reddy}]{behnamghader2024llm2vec}
Parishad BehnamGhader, Vaibhav Adlakha, Marius Mosbach, Dzmitry Bahdanau, Nicolas Chapados, and Siva Reddy. 2024.
\newblock \href {https://arxiv.org/abs/2404.05961} {{LLM2Vec}: Large language models are secretly powerful text encoders}.
\newblock \emph{arXiv preprint arXiv:2404.05961}.

\bibitem[{Berdichevsky et~al.(2025)Berdichevsky, Nahum-Gefen, and Ben~Zaken}]{berdichevsky2025salsa}
Ruslan Berdichevsky, Shai Nahum-Gefen, and Elad Ben~Zaken. 2025.
\newblock \href {https://arxiv.org/abs/2510.22691} {{SALSA}: Single-pass autoregressive {LLM} structured classification}.
\newblock \emph{arXiv preprint arXiv:2510.22691}.

\bibitem[{Brown et~al.(2020)Brown, Mann, Ryder, Subbiah, Kaplan, Dhariwal, Neelakantan, Shyam, Sastry, Askell et~al.}]{brown2020language}
Tom Brown, Benjamin Mann, Nick Ryder, Melanie Subbiah, Jared~D Kaplan, Prafulla Dhariwal, Arvind Neelakantan, Pranav Shyam, Girish Sastry, Amanda Askell, and 1 others. 2020.
\newblock Language models are few-shot learners.
\newblock \emph{Advances in neural information processing systems}, 33:1877--1901.

\bibitem[{Casanueva et~al.(2020)Casanueva, Tem{\v{c}}inas, Gerz, Henderson, and Vuli{\'c}}]{casanueva-etal-2020-efficient}
I{\~n}igo Casanueva, Tadas Tem{\v{c}}inas, Daniela Gerz, Matthew Henderson, and Ivan Vuli{\'c}. 2020.
\newblock \href {https://doi.org/10.18653/v1/2020.nlp4convai-1.5} {Efficient intent detection with dual sentence encoders}.
\newblock In \emph{Proceedings of the 2nd Workshop on Natural Language Processing for Conversational AI}, pages 38--45, Online. Association for Computational Linguistics.

\bibitem[{Chen et~al.(2019)Chen, Zhuo, and Wang}]{chen2019bert}
Qian Chen, Zhu Zhuo, and Wen Wang. 2019.
\newblock Bert for joint intent classification and slot filling.
\newblock \emph{arXiv preprint arXiv:1902.10909}.

\bibitem[{Devlin et~al.(2019)Devlin, Chang, Lee, and Toutanova}]{devlin2019bert}
Jacob Devlin, Ming-Wei Chang, Kenton Lee, and Kristina Toutanova. 2019.
\newblock Bert: Pre-training of deep bidirectional transformers for language understanding.
\newblock In \emph{Proceedings of the 2019 conference of the North American chapter of the association for computational linguistics: human language technologies, volume 1 (long and short papers)}, pages 4171--4186.

\bibitem[{Ding et~al.(2023)Ding, Qin, Yang, Wei, Yang, Su, Hu, Chen, Chan, Chen et~al.}]{ding2023parameter}
Ning Ding, Yujia Qin, Guang Yang, Fuchao Wei, Zonghan Yang, Yusheng Su, Shengding Hu, Yulin Chen, Chi-Min Chan, Weize Chen, and 1 others. 2023.
\newblock Parameter-efficient fine-tuning of large-scale pre-trained language models.
\newblock \emph{Nature machine intelligence}, 5(3):220--235.

\bibitem[{Dong et~al.(2024)Dong, Ruan, Cai, Lai, Xu, Pan, and Chen}]{dong2024xgrammar}
Yixin Dong, Charlie~F. Ruan, Yaxing Cai, Ruihang Lai, Ziyi Xu, Yilong Pan, and Tianqi Chen. 2024.
\newblock \href {https://arxiv.org/abs/2411.15100} {{XGrammar}: Flexible and efficient structured generation engine for large language models}.
\newblock \emph{arXiv preprint arXiv:2411.15100}.

\bibitem[{Enevoldsen et~al.(2025)Enevoldsen, Chung, Kerboua, Kardos, Mathur, Stap, Gala, Siblini, Krzemiński, Winata, Sturua, Utpala, Ciancone, Schaeffer, Sequeira, Misra, Dhakal, Rystrøm, Solomatin, Ömer Çağatan, Kundu, Bernstorff, Xiao, Sukhlecha, Pahwa, Poświata, GV, Ashraf, Auras, Plüster, Harries, Magne, Mohr, Hendriksen, Zhu, Gisserot-Boukhlef, Aarsen, Kostkan, Wojtasik, Lee, Šuppa, Zhang, Rocca, Hamdy, Michail, Yang, Faysse, Vatolin, Thakur, Dey, Vasani, Chitale, Tedeschi, Tai, Snegirev, Günther, Xia, Shi, Lù, Clive, Krishnakumar, Maksimova, Wehrli, Tikhonova, Panchal, Abramov, Ostendorff, Liu, Clematide, Miranda, Fenogenova, Song, Safi, Li, Borghini, Cassano, Su, Lin, Yen, Hansen, Hooker, Xiao, Adlakha, Weller, Reddy, and Muennighoff}]{enevoldsen2025mmtebmassivemultilingualtext}
Kenneth Enevoldsen, Isaac Chung, Imene Kerboua, Márton Kardos, Ashwin Mathur, David Stap, Jay Gala, Wissam Siblini, Dominik Krzemiński, Genta~Indra Winata, Saba Sturua, Saiteja Utpala, Mathieu Ciancone, Marion Schaeffer, Gabriel Sequeira, Diganta Misra, Shreeya Dhakal, Jonathan Rystrøm, Roman Solomatin, and 67 others. 2025.
\newblock \href {https://doi.org/10.48550/arXiv.2502.13595} {Mmteb: Massive multilingual text embedding benchmark}.
\newblock \emph{arXiv preprint arXiv:2502.13595}.

\bibitem[{Gao et~al.(2020)Gao, Fisch, and Chen}]{gao2020making}
Tianyu Gao, Adam Fisch, and Danqi Chen. 2020.
\newblock Making pre-trained language models better few-shot learners.
\newblock \emph{arXiv preprint arXiv:2012.15723}.

\bibitem[{Geng et~al.(2023)Geng, Josifoski, Peyrard, and West}]{geng2023grammar}
Saibo Geng, Martin Josifoski, Maxime Peyrard, and Robert West. 2023.
\newblock Grammar-constrained decoding for structured nlp tasks without finetuning.
\newblock \emph{arXiv preprint arXiv:2305.13971}.

\bibitem[{Goo et~al.(2018)Goo, Gao, Hsu, Huo, Chen, Hsu, and Chen}]{goo2018slot}
Chih-Wen Goo, Guang Gao, Yun-Kai Hsu, Chih-Li Huo, Tsung-Chieh Chen, Keng-Wei Hsu, and Yun-Nung Chen. 2018.
\newblock Slot-gated modeling for joint slot filling and intent prediction.
\newblock In \emph{Proceedings of the 2018 Conference of the North American Chapter of the Association for Computational Linguistics: Human Language Technologies, Volume 2 (Short Papers)}, pages 753--757.

\bibitem[{He et~al.(2020)He, Liu, Gao, and Chen}]{he2020deberta}
Pengcheng He, Xiaodong Liu, Jianfeng Gao, and Weizhu Chen. 2020.
\newblock Deberta: Decoding-enhanced bert with disentangled attention.
\newblock \emph{arXiv preprint arXiv:2006.03654}.

\bibitem[{Kwon et~al.(2023)Kwon, Li, Zhuang, Sheng, Zheng, Yu, Gonzalez, Zhang, and Stoica}]{kwon2023efficient}
Woosuk Kwon, Zhuohan Li, Siyuan Zhuang, Ying Sheng, Lianmin Zheng, Cody~Hao Yu, Joseph Gonzalez, Hao Zhang, and Ion Stoica. 2023.
\newblock Efficient memory management for large language model serving with pagedattention.
\newblock In \emph{Proceedings of the 29th symposium on operating systems principles}, pages 611--626.

\bibitem[{Lehmann et~al.(2015)Lehmann, Isele, Jakob, Jentzsch, Kontokostas, Mendes, Hellmann, Morsey, Van~Kleef, Auer et~al.}]{lehmann2015dbpedia}
Jens Lehmann, Robert Isele, Max Jakob, Anja Jentzsch, Dimitris Kontokostas, Pablo~N Mendes, Sebastian Hellmann, Mohamed Morsey, Patrick Van~Kleef, S{\"o}ren Auer, and 1 others. 2015.
\newblock Dbpedia--a large-scale, multilingual knowledge base extracted from wikipedia.
\newblock \emph{Semantic web}, 6(2):167--195.

\bibitem[{Lester et~al.(2021)Lester, Al-Rfou, and Constant}]{lester2021prompt}
Brian Lester, Rami Al-Rfou, and Noah Constant. 2021.
\newblock The power of scale for parameter-efficient prompt tuning.
\newblock In \emph{Proceedings of the 2021 Conference on Empirical Methods in Natural Language Processing}, pages 3045--3059.

\bibitem[{Li et~al.(2021)Li, Arora, Chen, Gupta, Gupta, and Mehdad}]{li2021mtop}
Haoran Li, Abhinav Arora, Shuohui Chen, Anchit Gupta, Sonal Gupta, and Yashar Mehdad. 2021.
\newblock {MTOP}: A comprehensive multilingual task-oriented semantic parsing benchmark.
\newblock In \emph{Proceedings of the 16th Conference of the European Chapter of the Association for Computational Linguistics: Main Volume}, pages 2950--2962.

\bibitem[{Li and Liang(2021)}]{li2021prefix}
Xiang~Lisa Li and Percy Liang. 2021.
\newblock Prefix-tuning: Optimizing continuous prompts for generation.
\newblock \emph{arXiv preprint arXiv:2101.00190}.

\bibitem[{Liu et~al.(2019)Liu, Ott, Goyal, Du, Joshi, Chen, Levy, Lewis, Zettlemoyer, and Stoyanov}]{liu2019roberta}
Yinhan Liu, Myle Ott, Naman Goyal, Jingfei Du, Mandar Joshi, Danqi Chen, Omer Levy, Mike Lewis, Luke Zettlemoyer, and Veselin Stoyanov. 2019.
\newblock Roberta: A robustly optimized bert pretraining approach.
\newblock \emph{arXiv preprint arXiv:1907.11692}.

\bibitem[{Ma et~al.(2025)Ma, Chochlakis, Pandiyan, Thomason, and Narayanan}]{ma2025multilabel}
Marcus Ma, Georgios Chochlakis, Niyantha~Maruthu Pandiyan, Jesse Thomason, and Shrikanth Narayanan. 2025.
\newblock \href {https://arxiv.org/abs/2505.17510} {Large language models do multi-label classification differently}.
\newblock \emph{arXiv preprint arXiv:2505.17510}.

\bibitem[{McAuley and Leskovec(2013)}]{mcauley2013hidden}
Julian McAuley and Jure Leskovec. 2013.
\newblock Hidden factors and hidden topics: understanding rating dimensions with review text.
\newblock In \emph{Proceedings of the 7th ACM conference on Recommender systems}, pages 165--172.

\bibitem[{Muennighoff et~al.(2022)Muennighoff, Tazi, Magne, and Reimers}]{muennighoff2022mteb}
Niklas Muennighoff, Nouamane Tazi, Lo{\"\i}c Magne, and Nils Reimers. 2022.
\newblock \href {https://doi.org/10.48550/ARXIV.2210.07316} {Mteb: Massive text embedding benchmark}.
\newblock \emph{arXiv preprint arXiv:2210.07316}.

\bibitem[{Ouyang et~al.(2022)Ouyang, Wu, Jiang, Almeida, Wainwright, Mishkin, Zhang, Agarwal, Slama, Ray et~al.}]{ouyang2022training}
Long Ouyang, Jeffrey Wu, Xu~Jiang, Diogo Almeida, Carroll Wainwright, Pamela Mishkin, Chong Zhang, Sandhini Agarwal, Katarina Slama, Alex Ray, and 1 others. 2022.
\newblock Training language models to follow instructions with human feedback.
\newblock \emph{Advances in neural information processing systems}, 35:27730--27744.

\bibitem[{Pang et~al.(2008)Pang, Lee et~al.}]{pang2008opinion}
Bo~Pang, Lillian Lee, and 1 others. 2008.
\newblock Opinion mining and sentiment analysis.
\newblock \emph{Foundations and Trends{\textregistered} in information retrieval}, 2(1--2):1--135.

\bibitem[{Rahman et~al.(2020)Rahman, Hasan, Lee, Zadeh, Mao, Morency, and Hoque}]{rahman2020integrating}
Wasifur Rahman, Md~Kamrul Hasan, Sangwu Lee, Amir Zadeh, Chengfeng Mao, Louis-Philippe Morency, and Ehsan Hoque. 2020.
\newblock Integrating multimodal information in large pretrained transformers.
\newblock In \emph{Proceedings of the conference. Association for computational linguistics. Meeting}, volume 2020, page 2359.

\bibitem[{Ren et~al.(2026)Ren, Zewde, Shen, Zhou, Ng, Raj, Duong, Zhang, and Tiangratanakul}]{ren2026openjev}
Simiao Ren, Kidus Zewde, Xingyu Shen, Yuchen Zhou, Dennis Ng, Ankit Raj, Tommy Duong, Yuxin Zhang, and Neo Tiangratanakul. 2026.
\newblock \href {https://arxiv.org/abs/2609.23959} {Open-jev judgments on {CallScreenBench}: Calibrated one-pass scam screening with a small language model}.
\newblock \emph{arXiv preprint arXiv:2609.23959}.

\bibitem[{Schick and Sch{\"u}tze(2020)}]{schick2020exploiting}
Timo Schick and Hinrich Sch{\"u}tze. 2020.
\newblock Exploiting cloze questions for few shot text classification and natural language inference.
\newblock \emph{arXiv preprint arXiv:2001.07676}.

\bibitem[{Socher et~al.(2013)Socher, Perelygin, Wu, Chuang, Manning, Ng, and Potts}]{socher2013recursive}
Richard Socher, Alex Perelygin, Jean Wu, Jason Chuang, Christopher~D Manning, Andrew~Y Ng, and Christopher Potts. 2013.
\newblock Recursive deep models for semantic compositionality over a sentiment treebank.
\newblock In \emph{Proceedings of the 2013 conference on empirical methods in natural language processing}, pages 1631--1642.

\bibitem[{Sun et~al.(2023)Sun, Li, Li, Wu, Guo, Zhang, and Wang}]{sun2023text}
Xiaofei Sun, Xiaoya Li, Jiwei Li, Fei Wu, Shangwei Guo, Tianwei Zhang, and Guoyin Wang. 2023.
\newblock Text classification via large language models.
\newblock \emph{arXiv preprint arXiv:2305.08377}.

\bibitem[{Tam et~al.(2024)Tam, Wu, Tsai, Lin, Lee, and Chen}]{tam2024letmespeak}
Zhi~Rui Tam, Cheng-Kuang Wu, Yi-Lin Tsai, Chieh-Yen Lin, Hung-yi Lee, and Yun-Nung Chen. 2024.
\newblock \href {https://arxiv.org/abs/2408.02442} {Let me speak freely? a study on the impact of format restrictions on performance of large language models}.
\newblock \emph{arXiv preprint arXiv:2408.02442}.

\bibitem[{Tsai et~al.(2019)Tsai, Bai, Liang, Kolter, Morency, and Salakhutdinov}]{tsai2019multimodal}
Yao-Hung~Hubert Tsai, Shaojie Bai, Paul~Pu Liang, J~Zico Kolter, Louis-Philippe Morency, and Ruslan Salakhutdinov. 2019.
\newblock Multimodal transformer for unaligned multimodal language sequences.
\newblock In \emph{Proceedings of the conference. Association for computational linguistics. Meeting}, volume 2019, page 6558.

\bibitem[{{TypeSafe AI}(2026)}]{typesafe2026systemone}
{TypeSafe AI}. 2026.
\newblock \href {https://typesafe.ai/blog/introducing-system-one-models-and-jev} {Introducing system one models and {Jev}}.
\newblock TypeSafe AI Blog.

\bibitem[{Wang et~al.(2024)Wang, Bai, Tan, Wang, Fan, Bai, Chen, Liu, Wang, Ge et~al.}]{wang2024qwen2}
Peng Wang, Shuai Bai, Sinan Tan, Shijie Wang, Zhihao Fan, Jinze Bai, Keqin Chen, Xuejing Liu, Jialin Wang, Wenbin Ge, and 1 others. 2024.
\newblock Qwen2-vl: Enhancing vision-language model's perception of the world at any resolution.
\newblock \emph{arXiv preprint arXiv:2409.12191}.

\bibitem[{Wei et~al.(2021)Wei, Bosma, Zhao, Guu, Yu, Lester, Du, Dai, and Le}]{wei2021finetuned}
Jason Wei, Maarten Bosma, Vincent~Y Zhao, Kelvin Guu, Adams~Wei Yu, Brian Lester, Nan Du, Andrew~M Dai, and Quoc~V Le. 2021.
\newblock Finetuned language models are zero-shot learners.
\newblock \emph{arXiv preprint arXiv:2109.01652}.

\bibitem[{Willard and Louf(2023)}]{willard2023outlines}
Brandon~T. Willard and R{\'e}mi Louf. 2023.
\newblock \href {https://arxiv.org/abs/2307.09702} {Efficient guided generation for large language models}.
\newblock \emph{arXiv preprint arXiv:2307.09702}.

\bibitem[{Yousefiramandi and Cooney(2025)}]{yousefiramandi2025finetuning}
Amirhossein Yousefiramandi and Ciaran Cooney. 2025.
\newblock \href {https://arxiv.org/abs/2512.12677} {Fine-tuning causal {LLMs} for text classification: Embedding-based vs. instruction-based approaches}.
\newblock \emph{arXiv preprint arXiv:2512.12677}.

\bibitem[{Zhang et~al.(2024)Zhang, Wang, Xu, Zhou, Gao, Su, Li, Chen et~al.}]{zhang2024mintrec2}
Hanlei Zhang, Xin Wang, Hua Xu, Qianrui Zhou, Kai Gao, Jianhua Su, Wenrui Li, Yanting Chen, and 1 others. 2024.
\newblock Mintrec2. 0: A large-scale benchmark dataset for multimodal intent recognition and out-of-scope detection in conversations.
\newblock \emph{arXiv preprint arXiv:2403.10943}.

\bibitem[{Zhang et~al.(2015)Zhang, Zhao, and LeCun}]{zhang2015character}
Xiang Zhang, Junbo Zhao, and Yann LeCun. 2015.
\newblock Character-level convolutional networks for text classification.
\newblock \emph{Advances in neural information processing systems}, 28.

\bibitem[{Zheng et~al.(2023)Zheng, Yin, Xie, Sun, Huang, Yu, Cao, Kozyrakis, Stoica, Gonzalez, Barrett, and Sheng}]{zheng2023sglang}
Lianmin Zheng, Liangsheng Yin, Zhiqiang Xie, Chuyue Sun, Jeff Huang, Cody~Hao Yu, Shiyi Cao, Christos Kozyrakis, Ion Stoica, Joseph~E. Gonzalez, Clark Barrett, and Ying Sheng. 2023.
\newblock \href {https://arxiv.org/abs/2312.07104} {{SGLang}: Efficient execution of structured language model programs}.
\newblock \emph{arXiv preprint arXiv:2312.07104}.

\end{thebibliography}

\appendix

\newpage
\centerline{\textbf{SUMMARY OF THE APPENDIX}}
\label{sec:appendix}
This appendix contains additional details for the \textbf{\textit{``Turning Generative LLMs into Low-Latency and Consistent Classifiers''}}. This is organized as follows:

\begin{itemize}
    \item \S\ref{app:additional_benchmarks:latency} \textbf{Serving Latency on SST-2}

    \item \S\ref{app:additional_benchmarks} \textbf{Additional Benchmarks}

    \item \S\ref{app:ablation_study} \textbf{Ablation Study}

    \item \S\ref{app:sensitivity_analysis} \textbf{Robustness and Generalization Analysis}
    \begin{itemize}
        \item\ref{app:sensitivity_analysis:label}~Effect of Label-Set Sizes
        \item\ref{app:sensitivity_analysis:batch}~Effect of Batch Size
        \item\ref{app:sensitivity_analysis:zero-shot}~Control-Token Permutations
        \item\ref{app:sensitivity_analysis:scale}~Scaling Analysis
    \end{itemize}

    \item \S\ref{app:data} \textbf{Data Construction}
    \begin{itemize}
        \item \ref{app:data:dataset_details}~Dataset Details
        \item \ref{app:data:implement_details}~Implement Details
    \end{itemize}

    \item \S\ref{app:preliminary_experiment} \textbf{Preliminary Study}
    \begin{itemize}
        \item\ref{app:preliminary_experiment:alt_prompt}~Alternative Prompting Baselines
    \end{itemize}

    \item \S\ref{app:prompt_configuration} \textbf{Prompt Configuration}

\end{itemize}

\section{Serving Latency on SST-2}
\label{app:additional_benchmarks:latency}
\begin{figure}[t]
    \centering
    \includegraphics[width=\columnwidth]{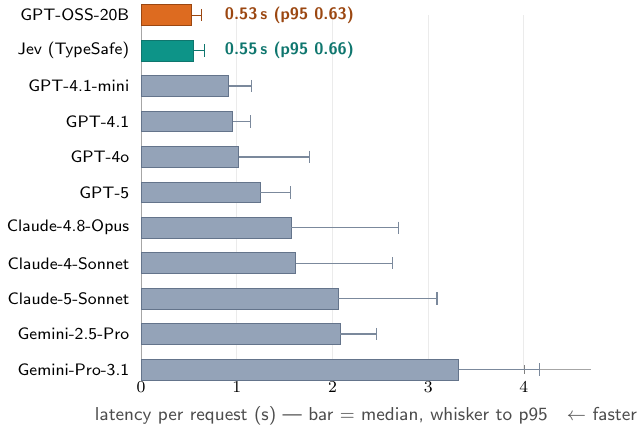}
    \caption{End-to-end per-request serving latency on SST-2 (872 examples), median (p50)
    with a whisker to the p95 tail, measured under the same round-robin, concurrency-1,
    shared serving-interface protocol as Figure~\ref{fig:text_latency_p50}. Orange = our \name{}
    single-token classifier; teal = the external single-token system Jev/TypeSafe; slate =
    frontier LLMs. The two single-token systems (\name{} 0.53\,s, Jev 0.55\,s) are tied at
    the front and well ahead of every frontier model, isolating the speedup to the
    single-token formulation itself.}
    \label{fig:sst2_latency}
\end{figure}

Figure~\ref{fig:sst2_latency} reports the same round-robin,
single-concurrency, shared serving-interface latency protocol used for the intent-routing set in
the main paper (Figure~\ref{fig:text_latency_p50}), here on the public SST-2 benchmark
(872 examples); the companion voice endpointing measurement is shown in the main text
(Figure~\ref{fig:voice_latency}).
The picture is identical to the intent-routing and voice settings: \name{} serves at 0.53\,s
median / 0.63\,s P95 on SST-2, essentially unchanged from its intent-routing latency despite the
very different task, and the external single-token system Jev/TypeSafe is essentially tied
(0.55\,s / 0.66\,s). Every frontier LLM is markedly slower and carries a wider P95 tail: they run
from GPT-4.1-mini at 0.92\,s median up to Gemini-Pro-3.1 at 3.32\,s. This confirms that
\name's latency advantage is a property of single-token decoding rather than of any particular
dataset, and that it holds on public data where results can be independently reproduced.

\section{Additional Benchmarks}
\label{app:additional_benchmarks}
\label{app:additional_benchmarks:other}
To complement the classical text benchmarks used in the main paper (SST-2, Amazon Reviews), we further evaluate \name on two modern and widely-used datasets~\citep{muennighoff2022mteb, enevoldsen2025mmtebmassivemultilingualtext}: \textbf{TweetTopic} (6 classes)~\citep{antypas2022twitter} and \textbf{Banking77} (77 classes)~\citep{casanueva-etal-2020-efficient}. These datasets are substantially more challenging. To show that the approach is not tied to a single backbone, we report \name applied to three different base models---Mistral-3-24B, GPT-OSS-20B, and Gemma-4-26B. As shown in
Table~\ref{tab:tweettopic-banking77}, every \name variant is competitive with GPT-4o while preserving the
single-token decoding advantage that keeps its latency well below decoder-based generation or GPT-4o; on the 77-class Banking77 task the GPT-OSS-20B and Gemma-4-26B backbones both surpass GPT-4o. These results further
highlight the practicality of \name for latency-sensitive text classification.

\begin{table*}[h!]
\centering
\vspace{0.5em}
\begin{tabular}{l l c}
\toprule
\textbf{Dataset} & \textbf{Model} & \textbf{Acc. (\%)} \\
\midrule
\multirow{5}{*}{\textbf{TweetTopic}}
  & GPT-4o                    & 77.50 \\
  & Mistral-3-24B (Base)      & 78.13 \\
  & \textbf{Mistral-3-24B (FT, \name)} & \textbf{81.86} \\
  & GPT-OSS-20B (FT, \name)   & 77.42 \\
  & Gemma-4-26B (FT, \name)   & 79.43 \\
\midrule
\multirow{5}{*}{\textbf{Banking77}}
  & GPT-4o                    & 74.68 \\
  & Mistral-3-24B (Base)      & 74.74 \\
  & Mistral-3-24B (FT, \name) & 70.94 \\
  & \textbf{GPT-OSS-20B (FT, \name)} & \textbf{79.21} \\
  & Gemma-4-26B (FT, \name)   & 77.28 \\
\bottomrule
\end{tabular}
\caption{Evaluation results on TweetTopic and Banking77 across three \name{} backbones (Mistral-3-24B, GPT-OSS-20B, Gemma-4-26B). All \name{} variants are competitive with GPT-4o while retaining single-token decoding; on Banking77 the GPT-OSS-20B and Gemma-4-26B backbones both surpass GPT-4o.}
\label{tab:tweettopic-banking77}
\end{table*}

\paragraph{Multilingual intent classification (MTOP).}
To strengthen our evaluation beyond English-only settings, we further evaluate \name on MTOP~\citep{li2021mtop}, a widely used multilingual intent-classification benchmark covering six languages (English, German, Spanish, French, Hindi, Thai). The results are in Table~\ref{tab:mtop}.

\begin{table*}[h!]
\centering
\small
\setlength{\tabcolsep}{3pt}  
\renewcommand{\arraystretch}{1.1}
\begin{tabular}{l|c|c|c|c|c|c|c}
\toprule
Model & en & de & es & fr & hi & th & Avg \\
\midrule
GPT-4o & 79.03 & 78.26 & 77.02 & 79.22 & 74.39 & 78.28 & 77.70 \\
Mistral-3-24B (Base) & 81.08 & 79.29 & 81.65 & 80.18 & 77.45 & 70.81 & 78.41 \\
Mistral-3-24B (FT, \name)
& 70.63 & 69.71 & 74.05 & 71.12 & 67.55 & 59.31 & 68.73 \\
GPT-OSS-20B (FT, \name)
& 71.99 & 70.62 & 71.84 & 71.73 & 68.90 & 68.46 & 70.59 \\
Gemma-4-26B (FT, \name)
& 75.06 & 70.75 & 73.95 & 76.45 & 73.07 & 72.73 & 73.67 \\
\bottomrule
\end{tabular}
\caption{Multilingual evaluation on the MTOP intent classification benchmark (6 languages) across three \name{} backbones. }
\label{tab:mtop}
\end{table*}

Since our fine-tuning corpus is entirely English, \name naturally achieves lower multilingual accuracy than GPT-4o and the multilingual Mistral base model. The gap is backbone-dependent: the Gemma-4-26B and GPT-OSS-20B backbones (average 73.7 and 70.6) generalize noticeably better across languages than the Mistral-3-24B fine-tune (68.7), and they avoid its sharp drop on Thai (59.3 vs.\ 68--73), indicating that the low-resource degradation reflects the English-only supervision rather than the single-token formulation itself. Across all backbones \name retains its key advantage—single-token decoding preserves the low-latency inference of our other benchmarks—highlighting that our classification formulation scales efficiently even in multilingual settings. We expect multilingual accuracy to improve substantially with multilingual supervision, which we identify as an important direction for future work.

\section{Ablation Study}
\label{app:ablation_study}
To assess the value of incorporating both unimodal and multimodal supervision,
we evaluated our fine-tuned Gemma-3-27B model on the held-out intent-routing dataset.
When adapted only on multimodal data, the model achieved 80.6\% accuracy.
By contrast, when adaptation combined both text-only
and multimodal data, accuracy improved to 84.7\% at the same single-token
decoding cost, i.e.\ without any latency penalty.
These results suggest that exposing the model to both text and multimodal supervision during adaptation
provides stronger representations and leads to consistent accuracy gains. A similar improvement ($\approx$0.5\%) at unchanged latency was also observed on DBpedia, indicating that the effect generalizes beyond a single dataset.

\section{Robustness and Generalization Analysis}
\label{app:sensitivity_analysis}

\subsection{Effect of Label-Set Sizes}
\label{app:sensitivity_analysis:label}
We further analyze the impact of the number of label sets on model performance by evaluating across datasets with increasing topic sizes, ranging from binary sentiment classification (SST-2, Amazon Reviews) to multi-class categorization (DBpedia with 14 topics).

\begin{figure*}[t]
    \centering
    \begin{subfigure}[t]{0.32\textwidth}
        \centering
        \includegraphics[width=\linewidth]{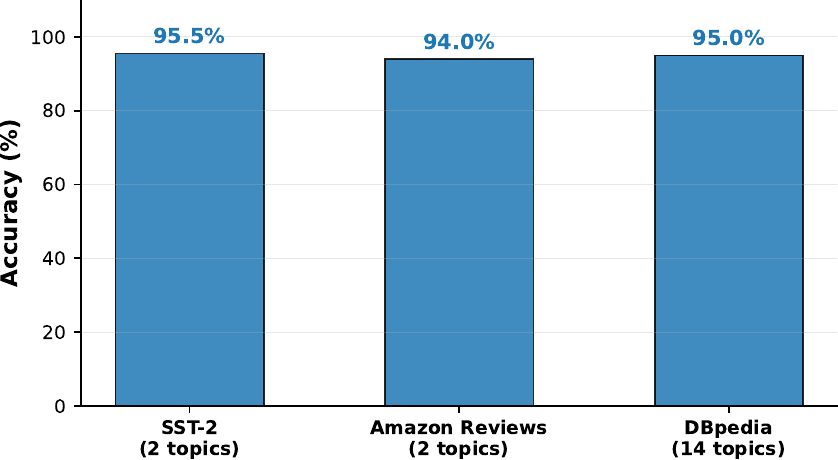}
        \caption{Topic cardinality.}
        \label{fig:topic_cardinality}
    \end{subfigure}
    \hspace{0.01\textwidth}
    \begin{subfigure}[t]{0.3\textwidth}
        \centering
        \includegraphics[width=\linewidth]{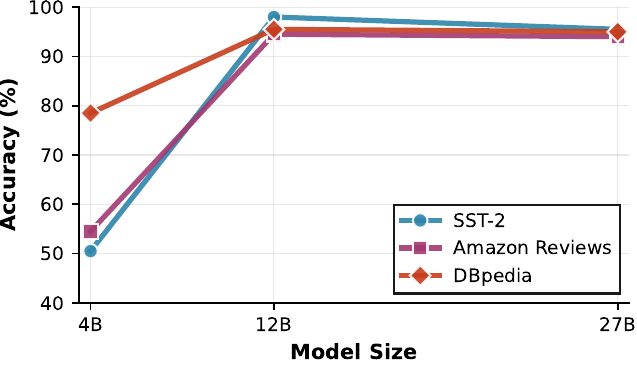}
        \caption{Accuracy vs model size.}
        \label{fig:accuracy_scaling}
    \end{subfigure}
    \hspace{0.01\textwidth}
    \begin{subfigure}[t]{0.3\textwidth}
        \centering
        \includegraphics[width=\linewidth]{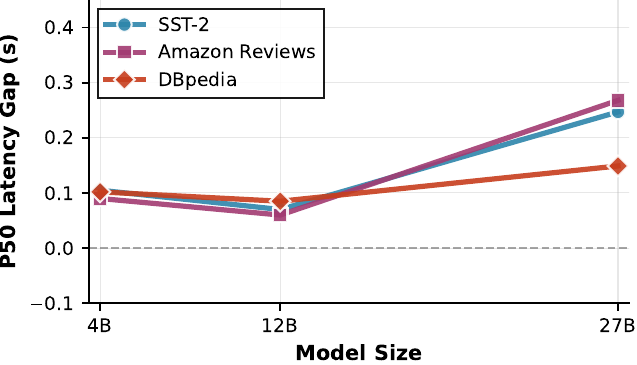}
        \caption{Latency imp. vs model size.}
        \label{fig:latency_scaling}
    \end{subfigure}
    \caption{Performance of Gemma-3 models across datasets.}
    \label{fig:combined_results}
\end{figure*}

\textbf{Generalization ability.} As shown in Figure~\ref{fig:topic_cardinality}, our fine-tuned Gemma-3-27B model consistently maintains high accuracy across datasets of varying difficulty. Even as the label space expands from 2 to 14 categories, the accuracy remains 95\% on DBpedia and comparable to binary sentiment datasets, demonstrating strong zero-shot generalization.

\textbf{Efficiency stability.} In addition to accuracy, we examine efficiency via P50 latency. The results reveal that latency remains essentially flat as the label space grows, since the model still emits a single control token regardless of how many labels are in play. This indicates that our design achieves scalable inference efficiency while handling tasks of increasing complexity.

\subsection{Effect of Batch Size}
\label{app:sensitivity_analysis:batch}
This section examines whether the latency benefits of \name persist under larger batch sizes, a
common setting in practical deployment scenarios (e.g., vLLM-based serving).  
While the main experiments fix the batch size to 1 for controlled comparison, real-world systems
frequently operate with significantly larger batches to maximize throughput. To assess robustness
under such conditions, we measure per-sample median (P50) and tail (P95) latencies for batch
sizes ranging from 2 to 64 on Amazon Reviews using Gemma-3-27B.

Across all batch sizes, \name consistently achieves substantially lower
latency than the base model. Notably, the improvement remains pronounced even in high-throughput
regimes (e.g., batch size 64), where inference engines are typically most optimized. These results
demonstrate that the single-token decision mechanism of \name yields durable efficiency gains that
are insensitive to batching.

\paragraph{Amazon Reviews.} Table~\ref{tab:batch_amazon} reports per-sample latency across
batch sizes. \name maintains a clear advantage over the base model.

\begin{table*}[h!]
\centering
\small
\begin{tabular}{rcccc}
\toprule
\textbf{Batch} &
\textbf{Base P50} &
\textbf{\name P50 (improv.)} &
\textbf{Base P95} &
\textbf{\name P95 (improv.)} \\
\midrule
2  & 0.1106 & 0.0545 (\textbf{0.0561$\downarrow$}) & 0.1175 & 0.0875 (\textbf{0.0300$\downarrow$}) \\
4  & 0.0554 & 0.0327 (\textbf{0.0227$\downarrow$}) & 0.0580 & 0.0427 (\textbf{0.0153$\downarrow$}) \\
8  & 0.0282 & 0.0205 (\textbf{0.0077$\downarrow$}) & 0.0295 & 0.0222 (\textbf{0.0073$\downarrow$}) \\
16 & 0.0146 & 0.0108 (\textbf{0.0038$\downarrow$}) & 0.0148 & 0.0110 (\textbf{0.0038$\downarrow$}) \\
32 & 0.0090 & 0.0068 (\textbf{0.0022$\downarrow$}) & 0.0098 & 0.0069 (\textbf{0.0029$\downarrow$}) \\
64 & 0.0269 & 0.0056 (\textbf{0.0213$\downarrow$}) & 0.0365 & 0.0053 (\textbf{0.0312$\downarrow$}) \\
\bottomrule
\end{tabular}
\caption{Amazon Reviews — Per-sample latency (seconds) vs.\ batch size for Gemma-3-27B.}
\label{tab:batch_amazon}
\end{table*}

We observe that per-sample latency generally decreases as batch size increases. Additionally, these results confirm that the latency improvements of \name are not limited to single-example inference, but extend reliably to the large-batch regimes commonly used for high-throughput deployment.

\subsection{Zero-Shot Adaptation via Control-Token Permutations}
\label{app:sensitivity_analysis:zero-shot}

This section examines the model's ability to generalize to unseen mappings between semantic labels and control tokens. During training, the correspondence between labels and control tokens is randomized across instances, preventing the model from relying on fixed token identities. To assess whether the model can operate under entirely new mappings at inference time, we evaluate the fine-tuned \name model on MIntRec~2.0 while applying random permutations of the control-token assignments.

In each of ten runs, a set of 30 control tokens is sampled from the 500-token pool and assigned to the 30 intent labels according to a fresh random permutation. The fine-tuned Gemma-3-27B model is then evaluated on the full test set (2{,}033 samples, with video) using this new mapping. This procedure ensures that every run tests the model under a previously unseen association between output tokens and semantic categories. Accuracy remains stable across permutations and on par with the fixed-mapping result: the mean accuracy is \textbf{61.62\%} with a standard deviation of \textbf{1.11}, close to the \textbf{62.72\%} reported for the same model in Table~\ref{tab:mintrec_results}. This indicates that the model interprets the label descriptions provided in the prompt rather than memorizing token identities. Table~\ref{tab:zero-shot-permutation} summarizes the results.

\begin{table}[h]
\centering
\begin{tabular}{c c c}
\toprule
\textbf{Run} & \textbf{Seed} & \textbf{Accuracy (\%)} \\
\midrule
1  & 42 & 61.93 \\
2  & 43 & 59.17 \\
3  & 44 & 60.90 \\
4  & 45 & 61.49 \\
5  & 46 & 63.40 \\
6  & 47 & 61.24 \\
7  & 48 & 61.98 \\
8  & 49 & 60.99 \\
9  & 50 & 62.76 \\
10 & 51 & 62.32 \\
\midrule
\multicolumn{2}{c}{Mean $\pm$ Std} & 61.62 $\pm$ 1.11 \\
\bottomrule
\end{tabular}
\caption{Accuracy under random permutations of control-token assignments during inference (Gemma-3-27B, full MIntRec~2.0 test set). Accuracy stays on par with the fixed-mapping result (62.72\%, Table~\ref{tab:mintrec_results}).}
\label{tab:zero-shot-permutation}
\end{table}

\paragraph{Procedure for swapping label sets at inference.}
Adapting \name to a new label set at inference time requires no retraining, adapter swapping, or model modification. The operator updates the system prompt to list the new label names and their natural-language descriptions, each mapped to a reserved control token (\texttt{[control\_1]}, \texttt{[control\_2]}, \dots), following the same template used during training (Appendix~\ref{app:prompt_configuration}). The decision space is then restricted to the control tokens for the active labels, and the model emits one of them in a single decode step, conditioning on the descriptions in the prompt rather than on fixed token identities. Because the mapping between labels and tokens is randomized during training, any assignment of labels to reserved tokens is valid at inference; the permutation results above confirm that accuracy is stable under such reassignments. This makes label additions, merges, and deprecations---common in production intent inventories---a prompt-side edit rather than a retraining event.

\subsection{Scaling Analysis}
\label{app:sensitivity_analysis:scale}
We further investigate how accuracy and efficiency scale with model sizes. 
\Cref{fig:accuracy_scaling,fig:latency_scaling} report results for Gemma-3 models with 4B, 12B, and 27B parameters on three text benchmarks. \textbf{Accuracy.} Performance improves consistently as model size increases (Figure~\ref{fig:combined_results}b). The 4B model shows lower accuracy, while the 12B model closes much of the gap. The 27B model achieves the strongest results across all datasets, exceeding 95\% on SST-2, Amazon Reviews, and DBpedia. \textbf{Latency.} Figure~\ref{fig:combined_results}c reports the \emph{P50 latency improvement} of \name{} relative to the corresponding base model. The improvement is positive across all datasets, showing that \name{} consistently reduces latency compared to standard autoregressive decoding. The improvement is modest for the 4B and 12B models, but becomes clearly larger for the 27B model. This trend indicates that \name{}’s efficiency advantage becomes more pronounced for larger models.

\section{Data Construction}
\label{app:data}

\subsection{Dataset Details}
\label{app:data:dataset_details}
We construct a balanced adaptation corpus with comparable amounts of text-only and multimodal classification instances, each pairing an input with a categorical label and spanning diverse intent- and topic-classification domains. All instances are reformulated into a unified JSON schema (see Figure~\ref{fig:dataset_example}), with consistent \texttt{messages} fields and an optional \texttt{image\_path}. To mitigate label memorization, we randomize control-token assignments, mapping class labels to tokens drawn from a reserved pool of up to 500 control tokens. The evaluation benchmarks reported in this paper are held out from this corpus.

\begin{figure}[ht]
    \centering
    \begin{tcolorbox}[
        title=\texttt{Example of a Classification Training Instance},
    ]
    \begin{flushleft}
    \small
    \texttt{
    \{\\
    \textcolor{darkred}{\textbf{"messages"}}: [\\
    \ \ \{ "role": "system", "content": "You are a classification expert. 
    Topics: [control\_x] Complain, [control\_y] Praise, [control\_z] Apologise ..." \},\\
    \ \ \{ "role": "user", "content": "\#\#\# USER CONVERSATION HERE \#\#\#" \},\\
    \ \ \{ "role": "assistant", "content": "[control\_x]" \}\\
    ],\\
    \textcolor{darkred}{\textbf{"image\_path"}}: "path/to/image.jpg",\\
    \}
    }
    \end{flushleft}
    \end{tcolorbox}
    \caption{Illustration of a fine-tuning training instance in our classification datasets. 
    Each sample includes the structured \texttt{messages} field and optional \texttt{image\_path}.}
    \label{fig:dataset_example}
\end{figure}

\subsection{Implement Details}
\label{app:data:implement_details}
We adapt each base model with supervised fine-tuning, keeping the token embeddings and LM head trainable so that the newly added control tokens can be learned. Training uses mixed precision (bfloat16) with gradient checkpointing on standard multi-GPU infrastructure, and early stopping on validation loss.

For LM-BFF, we use RoBERTa-base as the backbone, follow the standard prompt-based fine-tuning setup, and use 1 in-context demonstration with 16 training samples per class (consistent with the~\cite{gao2020making}). Template and verbalizer settings exactly match those reported in LM-BFF. As expected, LM-BFF performance is sensitive to the number of sampled demonstrations: using more samples improves accuracy but also increases latency substantially.

\paragraph{Baseline configurations and fairness.}
The baselines differ in the amount of task-specific adaptation, and we configure each following its intended usage. MAG-BERT and MulT are trained on the \emph{full} MIntRec~2.0 training set (the benchmark's standard protocol), while LM-BFF is run in its designed few-shot regime. For the BERT/RoBERTa linear heads we use 16-shot fine-tuning by design: these encoders require a task-specific head retrained from scratch whenever the label set changes, whereas \name reuses a single adapter and adapts by editing the prompt---so training the heads on the full dataset would compare a per-task, retrained classifier against a general-purpose model on precisely the axis where the two paradigms differ. The commercial APIs (GPT-4o, GPT-5) do not expose fine-tuning and serve as strong out-of-the-box references. We make these differences explicit so that the single-token formulation is not conflated with the effect of supervised fine-tuning alone.

\section{Preliminary Study}
\label{app:preliminary_experiment}
\subsection{Alternative Prompting Baselines}
\label{app:preliminary_experiment:alt_prompt}
To examine whether simple prompting strategies can emulate the single-symbol classification behavior of our method, we evaluate alternative formulations of the form $[x, \text{cls}, \text{cls-description}]$ using GPT-4o on the MIntRec2.0 benchmark. In this setting, the model is provided with the utterance, associated video frames, and the full mapping between the $30$ intent classes and their corresponding indices. The prompt explicitly instructs the model to output only the integer associated with the predicted intent. An example prompt is shown in Figure~\ref{fig:dataset_example_class_index}.

\begin{figure}[ht]
    \centering
    \begin{tcolorbox}[
        title=\texttt{Example of a Class-Index Prompting Instance (MIntRec2.0)},
    ]
    \begin{flushleft}
    \small
    \texttt{
    \{\\
    \textcolor{darkred}{\textbf{"messages"}}: [\\
    \ \ \{ "role": "system", "content": "You are an intent classification system. Respond with the number (0--29) corresponding to the intent. 
    Mapping: 0: Acknowledge, 1: Advise, ..., 28: Thank, 29: Warn. Output the number (0--29):" \},\\
    \ \ \{ "role": "user", "content": "Thank you" \},\\
    \ \ \{ "role": "assistant", "content": "28" \}\\
    ],\\
    \textcolor{darkred}{\textbf{"image\_path"}}: "data:image/jpeg;base64,...",\\
    \}
    }
    \end{flushleft}
    \end{tcolorbox}
    \caption{Illustration of the class-index prompting baseline evaluated on MIntRec2.0.}
    \label{fig:dataset_example_class_index}
\end{figure}

Despite the explicit numeric constraint, the model does not consistently produce a single-symbol output. In a non-trivial number of cases (16.67\%), GPT-4o generates additional text or multi-token strings. For example, when the expected prediction is \texttt{4}, the model sometimes outputs a full sentence such as ``The intent of the utterance...'', or formats the answer as ``3: Apologise'' rather than emitting a single integer. Furthermore, integers above nine are represented as multi-token sequences under the GPT-4o tokenizer, which inherently increases decoding latency and introduces substantial variability in response length. These behaviors prevent the model from operating as a deterministic one-token classifier and stand in contrast to the stable single-token decoding enabled by our learned control-token interface. This analysis indicates that prompting-based alternatives are insufficient to achieve low-latency, single-token inference.

In terms of classification accuracy on MIntRec2.0, the class-index prompt described above (without constrained decoding) yields 45.12\% accuracy, and adding constrained decoding over the 30 label tokens further increases it to 45.56\%. Thus, alternative prompting and constrained decoding slightly reduce formatting and parsing errors, but the gains are modest and do not close the substantial gap to \name (e.g., 62.72\% accuracy for Gemma-3-27B \name in Table~\ref{tab:mintrec_results}).

\section{Prompt Configuration}
\label{app:prompt_configuration}
We provide representative prompt templates used for evaluation across datasets, as shown in \Cref{fig:dataset_example_positive,fig:dataset_example_amazon,fig:dataset_example_dbpedia,fig:dataset_example_topic_thank}.
\begin{figure*}[t]
    \centering
    \begin{minipage}[t]{0.48\textwidth}
    \begin{tcolorbox}[
        title=\texttt{Example Prompt (SST-2)},
    ]
    \begin{flushleft}
    \small
    \texttt{
    \{\\
    \textcolor{darkred}{\textbf{"messages"}}: [\\
    \ \ \{ "role": "system", "content": "You are a classification expert. 
    Topics: [control\_1] Negative, [control\_2] Positive. 
    Based on the overall sentiment expressed in this review, respond with the relevant control token:" \},\\
    \ \ \{ "role": "user", "content": "it's a charming and often affecting journey." \},\\
    \ \ \{ "role": "assistant", "content": "[control\_2]" \}\\
    ],\\
    \textcolor{darkred}{\textbf{"image\_path"}}: "none",\\
    \}
    }
    \end{flushleft}
    \end{tcolorbox}
    \caption{Prompt template for SST-2 movie reviews.}
    \label{fig:dataset_example_positive}
    \end{minipage}\hfill
    \begin{minipage}[t]{0.48\textwidth}
    \begin{tcolorbox}[
        title=\texttt{Example Prompt (Amazon Reviews)},
    ]
    \begin{flushleft}
    \small
    \texttt{
    \{\\
    \textcolor{darkred}{\textbf{"messages"}}: [\\
    \ \ \{ "role": "system", "content": "You are a classification expert. 
    Topics: [control\_1] Negative, [control\_2] Positive. 
    Based on the overall sentiment expressed in this review, respond with the relevant control token:" \},\\
    \ \ \{ "role": "user", "content": "DVD Player crapped out after one year. I also began having the incorrect disc problems that I've read about on here. The VCR still works, but the DVD side is useless. I understand that DVD players sometimes just quit on you, but after not even one year? To me that's a sign of bad quality. I'm giving up JVC after this as well. I'm sticking to Sony or giving another brand a shot." \},\\
    \ \ \{ "role": "assistant", "content": "[control\_1]" \}\\
    ],\\
    \textcolor{darkred}{\textbf{"image\_path"}}: "none",\\
    \}
    }
    \end{flushleft}
    \end{tcolorbox}
    \caption{Illustration of an inference prompt from Amazon Reviews (Negative sentiment).}
    \label{fig:dataset_example_amazon}
    \end{minipage}
\end{figure*}

\begin{figure*}[t]
    \centering
    \begin{minipage}[t]{0.48\textwidth}
    \begin{tcolorbox}[
        title=\texttt{Example Prompt (DBpedia)},
    ]
    \begin{flushleft}
    \small
    \texttt{
    \{\\
    \textcolor{darkred}{\textbf{"messages"}}: [\\
    \ \ \{ "role": "system", "content": "You are a classification expert. 
    Topics: [control\_1] Company, [control\_2] EducationalInstitution, [control\_3] Artist, [control\_4] Athlete, [control\_5] OfficeHolder, [control\_6] MeanOfTransportation, [control\_7] Building, [control\_8] NaturalPlace, [control\_9] Village, [control\_10] Animal, [control\_11] Plant, [control\_12] Album, [control\_13] Film, [control\_14] WrittenWork. 
    Based on the content of this article, respond with the relevant control token:" \},\\
    \ \ \{ "role": "user", "content": "Pizza Port Brewing Company is a brewpub with five locations in Southern California: Solana Beach, two in Carlsbad (Downtown and Bressi Ranch), Ocean Beach and San Clemente. A former Pizza Port location in San Marcos spun out of Pizza Port in 2006 and is now an independent operation, the Port Brewing Company / Lost Abbey brewery. It has received multiple awards, including \"Small Brewpub of the Year\" for both 2003 and 2004 by the Great American Beer Festival and six awards for its beers at the World Beer Cup." \},\\
    \ \ \{ "role": "assistant", "content": "[control\_1]" \}\\
    ],\\
    \textcolor{darkred}{\textbf{"image\_path"}}: "none",\\
    \}
    }
    \end{flushleft}
    \end{tcolorbox}
    \caption{Illustration of an inference prompt from DBpedia (Company category).}
    \label{fig:dataset_example_dbpedia}
    \end{minipage}\hfill
    \begin{minipage}[t]{0.48\textwidth}
    \begin{tcolorbox}[
        title=\texttt{Example Prompt (Topic Classification)},
    ]
    \begin{flushleft}
    \small
    \texttt{
    \{\\
    \textcolor{darkred}{\textbf{"messages"}}: [\\
    \ \ \{ "role": "system", "content": "You are a topic classification expert. Before making a decision, carefully follow all the topic-specific instructions/descriptions. 
    Topics: [control\_1] Acknowledge, [control\_2] Advise, [control\_3] Agree, [control\_4] Apologise, [control\_5] Arrange, [control\_6] Ask for help, [control\_7] Asking for opinions, [control\_8] Care, [control\_9] Comfort, [control\_{10}] Complain, [control\_{11}] Confirm, [control\_{12}] Criticize, [control\_{13}] Doubt, [control\_{14}] Emphasize, [control\_{15}] Explain, [control\_{16}] Flaunt, [control\_{17}] Greet, [control\_{18}] Inform, [control\_{19}] Introduce, [control\_{20}] Invite, [control\_{21}] Joke, [control\_{22}] Leave, [control\_{23}] Oppose, [control\_{24}] Plan, [control\_{25}] Praise, [control\_{26}] Prevent, [control\_{27}] Refuse, [control\_{28}] Taunt, [control\_{29}] Thank, [control\_{30}] Warn. 
    Based on the above conversation, respond with the relevant topic ID:" \},\\
    \ \ \{ "role": "user", "content": "Thank you so much for your help! I really appreciate it." \},\\
    \ \ \{ "role": "assistant", "content": "[control\_29]" \}\\
    ],\\
    \textcolor{darkred}{\textbf{"image\_path"}}: "none",\\
    \}
    }
    \end{flushleft}
    \end{tcolorbox}
    \caption{Illustration of a prompt for topic classification (Thank intent).}
    \label{fig:dataset_example_topic_thank}
    \end{minipage}
\end{figure*}

\end{document}